# Cognitive-Motor Integration in Assessing Bimanual Motor Skills

Erim Yanik[1], Xavier Intes[2], Suvranu De[1]

[1] College of Engineering, Florida A&M University and The Florida State University, FL, USA
[2] Biomedical Engineering Department, Rensselaer Polytechnic Institute, NY, USA

**Accurate assessment of bimanual motor skills is essential across various professions, yet, traditional methods often rely on subjective assessments or focus solely on motor actions, overlooking the integral role of cognitive processes. This study introduces a novel approach by leveraging deep neural networks (DNNs) to analyze and integrate both cognitive decision-making and motor execution. We tested this methodology by assessing laparoscopic surgery skills within the Fundamentals of Laparoscopic Surgery program, which is a prerequisite for general surgery certification. Utilizing video capture of motor actions and non-invasive functional near-infrared spectroscopy (fNIRS) for measuring neural activations, our approach precisely classifies subjects by expertise level and predicts FLS behavioral performance scores, significantly surpassing traditional single-modality assessments.**

The assessment of bimanual motor skills is critical in high-stakes fields for credentialing[1,2] and training[2–6]. Traditional assessment methods that rely on direct observation[3,7] and subjective evaluations[2,8] often overlook the complexity of these skills, focusing primarily on visible motor actions and hand-eye coordination. This approach, while practical, neglects the deeper aspects of technical skills, especially in high-stakes fields where cognitive processing and decision-making are integral[9,10], failing to capture the full spectrum of the expertise required.

Deep Neural Networks (DNNs) have provided new pathways to assess complex bimanual motor skills in high-stakes fields with increased efficiency and objectivity[1]. Despite their promise, the focus has largely been on motor actions[2], analyzed through videos[11–19] and sensor-based kinematics[20–29], limited by the sparse representation of cognitive functions in available datasets. Few studies have delved into analyzing neural activations obtained through non-invasive optical neuroimaging[9,10], with the majority of these studies limited to detection[30–34] and mental workload evaluation[35–38], with a limited focus on skill assessment[10]. A significant gap remains in comparing these neural activations directly with motor actions, especially in critical fields such as surgery[39], underscoring the need for a broader evaluation framework.

In this study, we introduce a novel approach by conducting a direct statistical comparative analysis between neural activations and motor actions for assessing bimanual motor skills using DNNs. We explore the synergy of these modalities in multimodal analysis, applied to precision and cognitive-demanding tasks, particularly within the Fundamentals of Laparoscopic Surgery (FLS) program (Fig. 1). We employed self-supervised contrastive learning[40] to extract motor actions from the videos and non-invasive functional near-infrared spectroscopy (fNIRS)[10] for monitoring brain activations. Our convolutional DNN then predicts skill levels and direct performance scores. In each case, the models were also statistically compared for their explanability. Our goal is to establish a more accurate, comprehensive, and universally applicable method for evaluating bimanual motor skills by capturing both cognitive and motor dynamics.

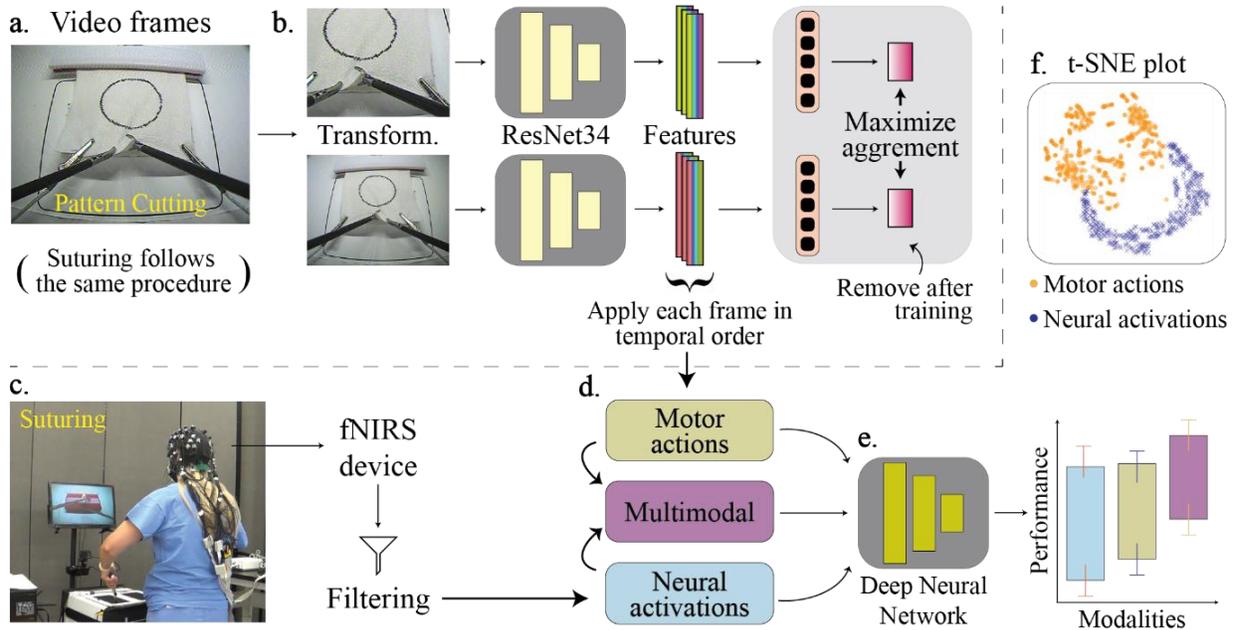

**Fig. 1 | Study overview. a.** Inputs include surgical video frames and non-invasive fNIRS neuroimaging. **b**. A self-supervised learning network (SimCLR) extracts motor actions from each frame, which are downscaled to align with the fNIRS channel data for pattern cutting and suturing tasks. **c.** fNIRS signals are collected simultaneously as videos and are postprocessed. **d.** Both modalities are independently analyzed and then combined for a comprehensive skill assessment. **e.** Deep Neural Network trains on this data, providing detailed skill assessment and performance metrics. **f.** t-SNE visualization demonstrates a distinct separation between neural activations (blue) and motor actions (orange).

## Results

**Dataset and analysis.** *Tasks*. We analyzed two aspects of the FLS program: pattern cutting (7 subjects / 475 trials) and suturing (18 subjects / 63 trials). Pattern cutting was evaluated using standard FLS scoring methods, with performance categorized based on successful execution[41]. Suturing skills were assessed using Objective structured assessment of technical skill (OSATS) scores and classified based on subject expertise levels, i.e., expert and novice.

*Analysis details*. First, we extracted self-supervised features from each video to generate spatiotemporal feature sets per video. Our DNN then underwent rigorous training and cross-validation (CV), i.e., leave-one-user-out (LOUO)[42], using these feature sets and corresponding scores and classes. We conducted 100 iterations for each scenario to ensure robustness against initialization bias, with results reflecting average performance across trials. Statistical significance was determined by comparing 100 runs of each modality via Student's t-test or Mann-Whitney U-test, based on normality.

**Assessment via neural activations.** *Score prediction*. Table 1 presents the performance metrics for assessing surgical skills. Our model demonstrated promising performance in predicting scores, achieving an $R^2$ of 0.889±0.011 for pattern cutting and 0.690±0.029 for suturing. Figure 2a visually compares these predicted scores with the actual, or ground-truth, performance scores.

*Classification.* For class predictions, the model's accuracies were 0.980±.006 and 1.0 in pattern cutting and suturing via neural activations. Further, we incorporated NetTrustScore (NTS)[43] to gauge the model's trustworthiness in correctly predicting each skill. The NTS leverages SoftMax of predictions to gauge the reliability of the results. The NTS values were 0.926±0.021 for the Fail class and 0.968±0.008 for the Pass class in pattern cutting. In suturing, the NTS values were 0.936±0.017 for the Resident class and 0.976±0.004 for the Surgeon class. Fig. 3a illustrates the trust spectrums[43].

| Table 1 \| Performance metrics with standard deviation (±) | | | | | |
|---|---|---|---|---|---|
| Dataset | Modality | R2 | Accuracy | NTS (Fail or Resident) | NTS (Pass or Surgeon) |
| Pattern cutting | NA | *0.889±.011* | *0.980±.006* | *0.926±.021* | *0.968±.008* |
|  | MA | 0.870±.011 | 0.976±.005 | 0.912±.012 | 0.961±.006 |
|  | MM | **0.897±.008** | **0.983±.005** | **0.936±.017** | **0.976±.004** |
| Suturing | NA | *0.690±.029* | 1.0 | 0.932±.015 | 0.926±.011 |
|  | MA | 0.662±.022 | 1.0 | *0.942±.008* | *0.932±.010* |
|  | MM | **0.729±.022** | N/A | **0.944±.010\*** | **0.946±.006** |

**Bold** formatting indicates the statistical significance (p < .05) of multimodality compared to neural activations (NN) and motor actions (MA). *Italic* denotes statistical significance (p < .05) of NA and MA against each other. *significance only against neural activations.

**Assessment via motor actions.** *Score prediction.* Our model demonstrated an $R^2$ of 0.870±.011 for pattern cutting and 0.662±.022 for suturing, respectively (Fig. 2b). Notably, neural activations resulted in significantly better score prediction than motor actions in both tasks (p<0.05), with a +2.2% increase in R2 for pattern cutting and +4.2% for suturing (see Table1).

*Classification.* We obtained accuracies of 0.976±.005 and perfect 1.0 for class predictions in pattern cutting and suturing. The performance in pattern cutting was marginally lower with motor actions compared to neural activations (a difference of -0.4%, p<0.05). The NTS values were 0.912±.012 and 0.961±.006 for pattern cutting and 0.942±.008 and 0.932±.010 for suturing in their respective unsatisfactory and satisfactory classes (Fig. 3b). Interestingly, motor actions showed a lower trustworthiness in pattern cutting but were higher in suturing (both p<0.05).

**Assessment via multimodality.** *Score prediction.* We combined motor actions and neural activations in a multimodality approach. This resulted in a score prediction performance ($R^2$) of 0.897±.008 for pattern cutting and 0.729±.022 for suturing (Fig. 2c). Compared to using motor actions alone, this dual-modality approach improved score prediction by 3.1% in pattern cutting and 10.1% in suturing (refer to Table 1).

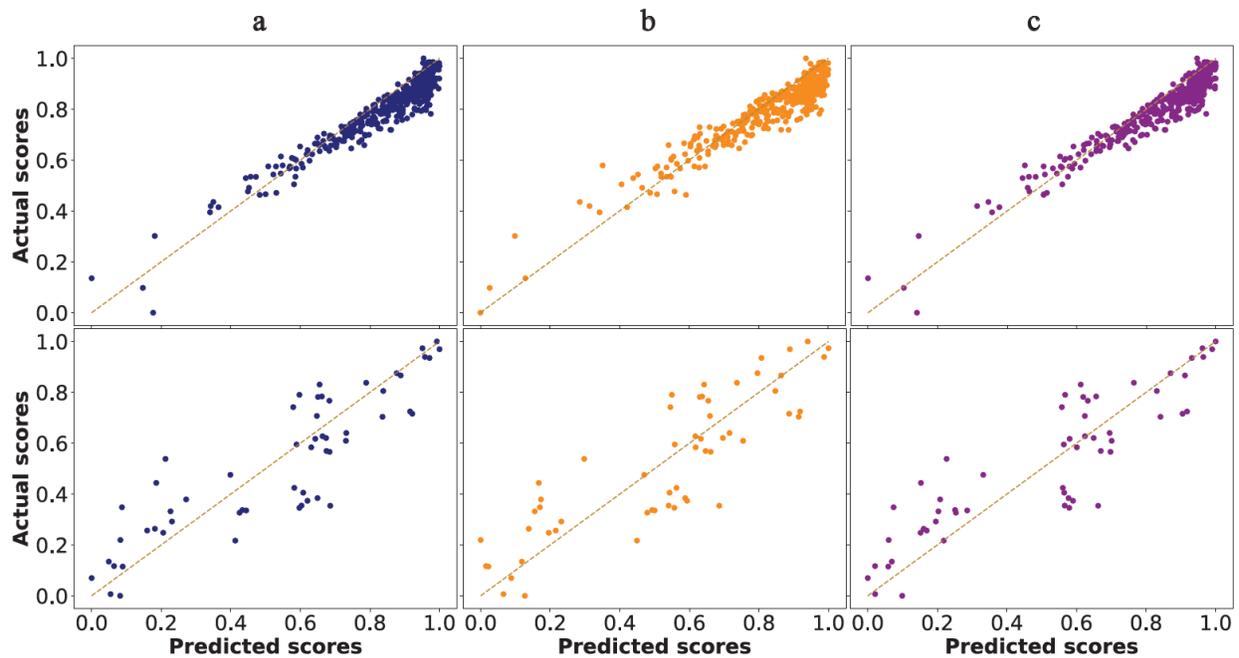

**Fig. 2 | Regression plots** for **a.** neural activations (blue), **b.** motor actions (orange), and **c.** multimodality (purple) for pattern cutting (top row) and suturing (bottom row).

*Classification.* For class prediction in pattern cutting, the accuracy reached 0.983±0.005, a significant enhancement over motor actions alone (0.7% | p<.05). In suturing, we did not perform a multimodal analysis due to perfect accuracies reported in individual modalities. The NTS values were 0.936±.017 and 0.976±.004 for pattern cutting while 0.944±.010 and 0.946±.006 for suturing in the respective classes (Fig. 3c). Multimodal input yielded greater reliability than each modality in isolation (p<.05).

**Discussion**
The current paradigm in evaluating bimanual motor skills is subjective and relies on manual observation of motor actions, a method that may not fully encapsulate the complexity of such tasks. Recent advancements in Deep Neural Networks (DNNs) have introduced automated and objective assessments[2]. However, these advancements have largely focused on motor actions as well, overlooking neural activations, despite evidence showing their ability to differentiate skills, particularly in high-stakes fields like surgery[39,44–46]. Our study bridges this gap by integrating cognitive aspects – neural activations – with motor actions for competency assessment using DNNs for the first time. This approach, tested via assessing surgical skills within the FLS program, not only effectively classifies skills and predicts performance scores for each modality but also highlights the importance of considering both complex cognitive functions and motor skills in high-stakes disciplines.

Our study's primary contribution is benchmarking neural activations against the traditional motor action modality. We found that neural activations offer robust performance in score prediction across both tasks (p<.05), vital for effective formative assessment and skill development[4,15]. Additionally, this modality demonstrates a comparable classification accuracy in pattern cutting compared to motor actions (p<.05 | Table 1), which is essential in the FLS

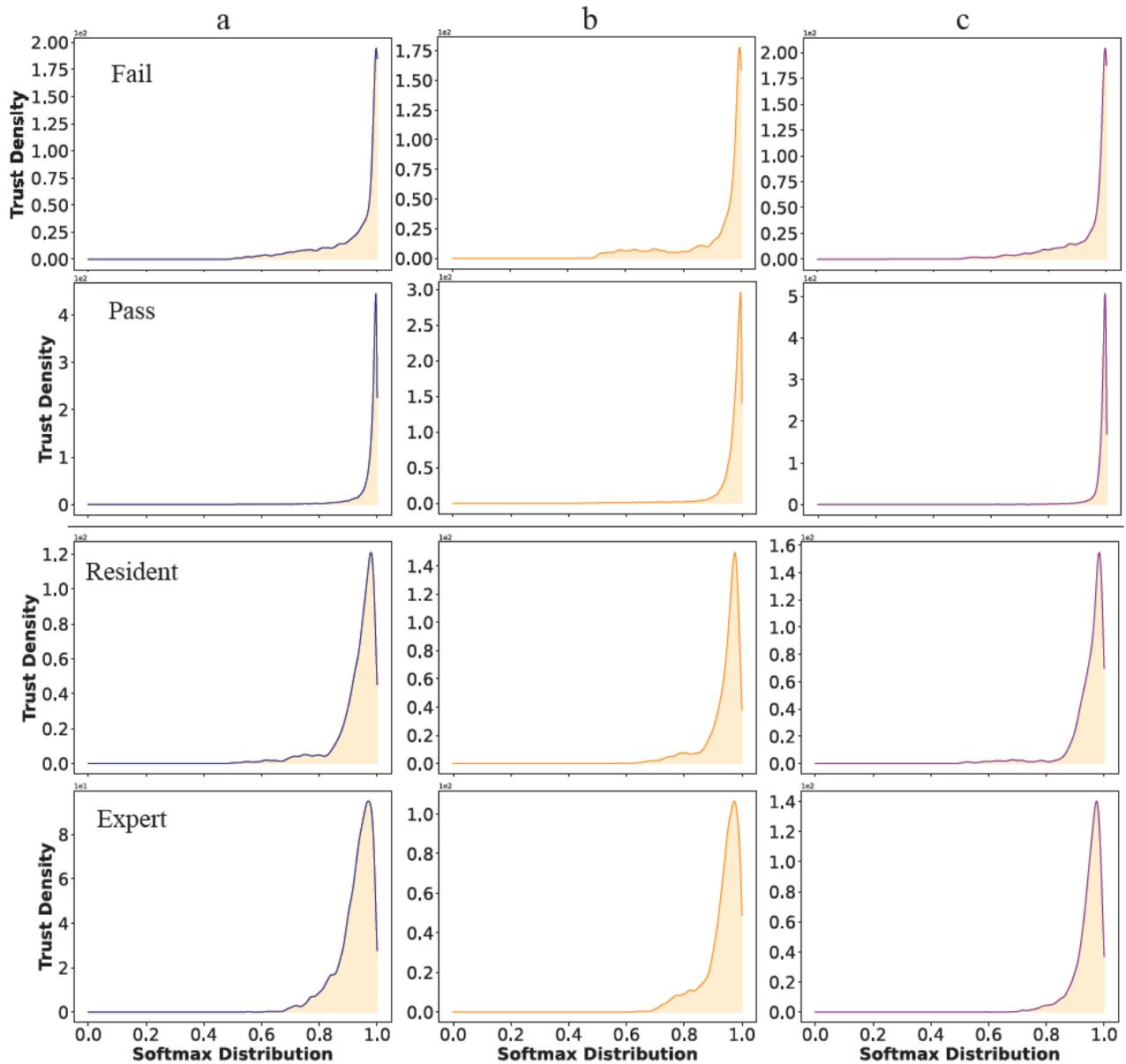

**Fig. 3 | NTS plots** for **a.** neural activations (blue), **b.** motor actions (orange), and **c.** multimodality (purple) in pattern cutting (above the center line) and suturing (below the center line).

certification process to uphold high standards of surgical competence[4,15]. These indicate that neural activations, a metric traditionally overlooked in favor of more observable motor actions, offer a valuable dimension to skill assessment and can be used as an alternative in skill-intensive fields.

The multimodal approach yielded a robust improvement in score prediction performance for both FLS tasks significantly improving the classification performance in pattern cutting (p<.05 | Table 1 | Fig. 2). These suggest that both modalities contribute uniquely, and the convergence of motor and neural data can provide a more nuanced and comprehensive understanding of surgical competency. Such an integrated approach could be pivotal in refining high-stakes training programs, where the focus extends beyond mere technical prowess to encompass the cognitive strategies that underpin expert performance. These hold promise for enhancing skill assessment in various domains where both physical dexterity and cognitive acumen are crucial for expertise.

Besides performance, a crucial aspect for deploying our DNN in high-stakes settings is its reliability, i.e., consistency across subjects whose data did not train the model. To evaluate this, we conducted a trustworthiness analysis for each modality based on true predictions, utilizing the NTS methodology[47,48]. As illustrated in Fig. 3, the trust density distributions for all modalities and tasks were skewed towards higher Softmax values, resulting in high NTS values (Table 1). This indicates that our model is capable of developing a classifier with a robust distinction between different skill levels. Consequently, it is reasonable to anticipate that the model would maintain its performance consistency across unseen subjects.

We now elucidate the rationale behind our selection of fNIRS and self-supervised features for representing neural activations and motor actions, respectively, in surgical contexts. A key consideration in surgical procedures is the freedom of motion, which is crucial for clinical outcomes[42]. Traditional neuroimaging methods[49], like electroencephalography (EEG) and functional Magnetic Resonance Imaging (fMRI), typically impose restrictions on movement. In contrast, fNIRS offers the advantage of allowing unhindered motion[46], a significant benefit in a surgical setting. Moreover, fNIRS provides superior spatial resolution compared to EEG and better temporal resolution than fMRI[50,51]. These attributes make fNIRS a more suitable choice for non-invasive optical neuroimaging in real-life surgical environments, with broader implications to similar bimanual skill-intensive fields.

Transitioning to motor action representation, we assessed the efficacy of self-supervised contrastive learning in extracting features from surgical videos by comparing it against established tool motion data[2,13,15,20–22,29,52] for the same pattern cutting dataset. As indicated in Table 2, both methods demonstrated comparable efficacy in evaluating surgical expertise. Despite their similar performance to tool motion data, the adoption of self-supervised features is justified by their inherent benefits. Primarily, self-supervised learning facilitates automated feature extraction, offering a more efficient and less labor-intensive alternative to the bounding box annotation required for tool motion analysis. Moreover, the dimensionality of the output feature sets in self-supervised contrastive learning can be adjusted. In contrast, tool motion data, reliant on sensor-based methods or labor-intensive video analysis, demands additional costly kinematic variables for similar expansion[2].

**Table 2 | Comparison for the self-supervised features (SS) and the tool motions (TM)**

| Input | Modality | R2 | Accuracy |
|---|---|---|---|
| SS | Motor Actions | 0.870±.011 | 0.976±.005 |
| TM |  | 0.899±.010 | 0.975±.005 |

We employed Class Activation Maps (CAMs)[2,15,21,22] to analyze how neural activations and motor actions concentrate on different sections of the tasks for skill assessment. Building upon our previous study[15], which extensively analyzed the statistical and visual aspects of CAMs in differentiating skill classes for the same pattern cutting dataset, this study expands on that foundation by statistically comparing motor actions with neural activations. We revealed distinct trends between these modalities by comparing CAM plots using the Spearman correlation coefficient (ρ). In pattern cutting, we found a weak correlation (ρ = .24) and a moderate correlation in suturing (ρ = .55) between neural activations and motor actions, suggesting each modality highlights different aspects of surgical skills. However, interpreting CAMs in multimodal cases is challenging for feedback, as discerning which parts of the CAM correspond to each modality is unclear. Consequently, we have refrained from providing an in-depth analysis of multimodality.

The limitation of our study is the constrained scope of our validation cohorts, which are currently limited to the context of the FLS program. To fully ascertain the applicability and effectiveness of our findings, there is a need to extend the validation to other skill-intensive fields. Such expansion would enable a more comprehensive comparison of neural activations and motor actions and provide deeper insights into the practical utility of multimodality in high-stakes environments.

Our study marks a significant advancement in bimanual skill assessment for high-stakes, skill-intensive fields by integrating cognitive and motor dimensions through DNNs. The distinct and complementary roles of neural activations and motor actions, as uncovered via surgical skill assessment, lay the groundwork for a more comprehensive and nuanced approach to evaluating skills in broad applications where professional training and competency are crucial.

## Methods

**Dataset characteristics.** *Tasks.* Our study utilized two laparoscopic tasks from the FLS program[53], recognized as a prerequisite for board certification in general surgery and ob/GYN surgery, with growing adoption in Canada, Australia, and Europe[53–55]. These tasks were laparoscopic pattern cutting and suturing. Our data collection adhered to the Institutional Review Board (IRB) guidelines of Rensselaer Polytechnic Institute and University at Buffalo, with informed consent obtained from all participants.

The laparoscopic pattern cutting task involved seven inexperienced residents (2 males and 5 females; average age 24 ± 0.82). They executed 475 trials over 12 consecutive days, cutting a circular pattern from gauze using laparoscopic tools[41]. Performance was evaluated using FLS scoring[53], based on time and precision[41]. Binary Pass and Fail class labels[41] were generated for each trial based on the FLS scores, resulting in 444 Pass and 31 Fail trials.

In the suturing task, 8 residents (5 males and 3 females) and 10 surgeons (5 males and 5 females), with an average age of 31 (std: 7.9), participated, producing 63 trials. The surgeons ranged from 1 to 20 years in FLS proficiency. This task involved suturing a Penrose drain, scored using the OSATS[56] criteria, focusing on time, deviation, incision gap, and knot security. One resident was excluded from score prediction due to consistently negative scores. The final dataset comprised 39 trials labeled as "Surgeon" and 24 as "Resident" based on the participants' expertise rather than OSATS Pass/Fail criteria due to limited Pass trials.

**Input modalities**. *Neural activations.* We used non-invasive fNIRS signals to represent neural activations. fNIRS uses near-infrared light through the scalp for the purpose of detecting the dissipated light by brain tissue[46]. Hence, fNIRS can monitor intracranial changes[57].

fNIRS signals were recorded, simultaneous to videos, using a continuous-wave near-infrared spectrometer: CW6, Techen Inc, at a sampling rate of 7.8125 Hz for pattern cutting. NIRSport2 and NIRx were used at a sample rate of 5.0863 Hz for suturing. Before the experimentation for each subject, we utilized the International 10-20 system to place the electrodes accurately. The preprocessing of fNIRS was done at the optical density level using the HomER3 toolbox[58]. First, it had baseline correction, artifact removal, e.g., systemic hemodynamic activations, and physiological noise filtering via bandpass filtering with 0.01 to 0.5 Hz. Next, three layers of motion correction were done using spline fitting to remove the remaining artifacts. Then, we converted the filtered data to changes in hemoglobin concentration using the Modified Beer-Lambert Law (MBLL)[59]. As a result, we generated the oxy-Haemoglobin concentration changes (ΔHbO) corresponding to six Prefrontal Cortex (PFC) channels for pattern cutting and eight for suturing. ΔHbO changes are linked to neural activations[46]. PFC is responsible for motor planning and decision making[57], and signals collected from this region have been used to differentiate between skill levels effectively[39].

*Motor actions.* Surgical videos represented motor actions. For the video collection, we utilized the built-in FLS box camera with a resolution of 640x480 and 720x480 for pattern cutting and suturing. The frame per second (FPS) rate was 30 for both.

A self-supervised contrastive learning model, SimCLR[40], was employed to extract salient spatiotemporal features from surgical videos. SimCLR enables us to take advantage of more pattern cutting videos (+1580) not used in this study due to a lack of non-invasive neuroimaging data. Moreover, it strengthens the model against blurry and jittery video frames. Further, self-supervised models do not require annotation, e.g., tool motion[15]. This saves time- and resources- as annotation is manual and relies on domain expertise.

SimCLR consisted of two steps. First, a backbone, $f_b(.) \in \mathbb{R}^D$, extracted D-dimensional features from the augmented versions of a given video frame. Here, $f_b(.)$ was ResNet34 pretrained on ImageNet. Then, a linear classifier, $f_h(.) \in \mathbb{R}^K$, processed these features to output K-(128-) dimensional features in hidden space. The model's objective was to maximize the likelihood of accurately detecting the augmented versions of the frame in the hidden space amongst a batch of uncorrelated frames. Finally, we removed $f_h(.)$ after the training was completed.

*Spatiotemporal feature generation.* We applied the trained $f_b(.)$ on each video frame in temporal order for a given trial. This generated the spatiotemporal features ($X_i$) for the trial, i.e., $X_i = [f_b(x_{i1}), ..., f_b(x_{ij}), ..., f_b(x_{iT_i})] \in \mathbb{R}^{T \times D}$. Here, $T_i$ is the temporal length of the trial $i$. Further, $x_i \in \mathbb{R}^{T_i \times H \times W \times 3}$ is the video frame list for a given trial $i$, where H and W are the height and width of each frame. In addition, $x_{ij}$ is the $j$<sup>th</sup> frame of the trial $i$. Finally, 1D GAP[60] was used to regulate the output feature space, i.e., $\mathbb{R}^{T_i \times D} \rightarrow \mathbb{R}^{T_i \times D'}$. Here, $D'$ is 6 and 8 for pattern cutting and suturing to match the number of features in fNIRS signals.

*Tool motion data.* In our previous study[15], the same pattern cutting dataset was used to extract tool motion data. Specifically, we used the detection model – Mask R-CNN[61] – to detect tools in each frame and merge them in temporal order to obtain spatiotemporal input. The input comprised the Cartesian coordinates for laparoscopic scissors and grasper, yielding a feature size of 4.

**Model development.** Our pipeline builds upon the Video-Based Assessment Network (VBA-Net)[15], detailed in our recent publication. In that study, tool motion data was derived from videos using an instance segmentation network, with an autoencoder then producing key spatiotemporal features for a classifier to assess surgical skills. In the current study, we adapted this approach by employing only the classifier component of the VBA-Net. For motor actions, the SimCLR framework replaces the instance segmentation network and autoencoder, efficiently extracting salient spatiotemporal features. Additionally, the fNIRS signals, being inherently spatiotemporal, were directly fed into the classifier.

The classifier's core involves the input ($X_i$) passing through a 1D convolutional layer, followed by a residual block containing two 1D convolutional layers and an identity shortcut. Spatial and channel squeeze and excitation (SCse)[62] attention is applied, leading into a 1D Global Average Pooling (GAP) and fully connected layers. The latter outputs either class probabilities for classification analysis or predicted performance scores for score prediction. Consistent with our previous work, the same hyperparameters as in the VBA-Net model[15] were utilized for this study.

**Training.** *SimCLR.* We developed distinct models for each cohort, utilizing them to extract spatiotemporal features from their respective trials. The frame split for training and validation was 143,287/17,373 in pattern cutting and 21,191/3,315 in suturing. Adhering to recommended augmentations[40], we applied random resized crop, jittering, Gaussian blur, horizontal flip, and grayscaling, normalizing all video frames.

The training was set for 200 epochs, incorporating an early stopping mechanism with the patience of 10 epochs – training ceased if there was no improvement over ten consecutive epochs, and the epoch with the lowest loss was saved. Given the performance boost with larger batch sizes[40] in self-supervised contrastive learning, batch sizes were set at 256 for pattern cutting and 512 for suturing.

*Classifier.* Spatiotemporal inputs were downsampled to 1 frame per second (FPS) for both tasks to balance computational efficiency and effectiveness[13]. fNIRS signals were similarly resampled for temporal alignment with the videos at 1 FPS, and min-max normalization was applied to each input. The model trained for up to 5000 epochs, again employing early stopping with a patience of 10 epochs. Due to variable temporal lengths, the batch size was set to one. We maintained consistent hyperparameters and training parameters across modalities, following the same procedure to compare the effectiveness of self-supervised features and tool motion data in representing motor actions.

**Evaluation.** The efficacy of self-supervised contrastive learning is gauged in the downstream task, i.e., DNN's performance on classification and score prediction. To assess the performance of the DNN, we employed the LOUO CV scheme. This approach is particularly relevant for certification programs where models are expected to evaluate new candidates who are tested infrequently. LOUO ensures that all trials from a single subject are excluded from training and used solely for validation, thereby measuring the model's effectiveness on unseen subjects.

For classification analysis, we measured the model's performance and reliability using accuracy and the NTS[43]. Score prediction analysis was conducted using the $R^2$ metric. Each scenario was run 100 times to ensure robustness, enabling statistical comparison across modalities. This process involved generating and analyzing the distributions of each metric after 100 iterations. Outliers were removed using Tukey Fences[63], and the Shapiro-Wilk test[64] was applied to assess normal distribution. Subsequently, we compared the metric distributions across modalities using either a one-sided Student's t-test or Mann-Whitney U-test[65], depending on distribution normality. Both tests were conducted with a significance threshold of 0.05, aiming to determine whether one modality outperformed the other in terms of mean distribution values.

Regarding our computational tools and resources, the self-supervised contrastive model was trained using Pytorch, while the DNN was developed in TensorFlow. Our computations were performed on the IBM Artificial Intelligence Multiprocessing Optimized System (AiMOS) at Rensselaer Polytechnic Institute, leveraging its 8 NVIDIA Tesla V100 GPUs, each with a capacity of 32 GB.

## Acknowledgments

The authors want to thank Dr. Yuanyuan Gao for pattern cutting data collection. Moreover, we acknowledge Dr. Lora Cavuoto for facilitating suturing experiments.

## Author contributions

E.Y., X.I., and S.D. conceived the idea. E.Y. developed the model pipeline, conducted the statistical and data analysis, and drafted the manuscript. X.I. and S.D. were responsible for supervising and revising the manuscript's content.

## Competing interests

The authors declare no competing interests.